\documentclass[letterpaper,10pt,conference]{ieeeconf}
\pdfoutput=1

\IEEEoverridecommandlockouts
\usepackage{graphics} 
\usepackage{epsfig} 
\usepackage{mathptmx} 
\usepackage{times} 
\usepackage{amsmath} 
\usepackage{amssymb} 
\usepackage{cite}
\usepackage{color}
\usepackage{epstopdf}
\usepackage{graphicx}
\usepackage{amsfonts}
\usepackage{mathtools}
\usepackage[mathcal]{euscript}
\usepackage{caption}
\usepackage[hidelinks]{hyperref}
\usepackage[T1]{fontenc} 
\usepackage[hyphenbreaks]{breakurl} 
\usepackage{mathdots} 

\usepackage{subcaption}
\DeclareCaptionLabelSeparator{periodspace}{.\quad}
\usepackage{amsthm}

\theoremstyle{definition} 
\newtheorem{example}{Example}

\title{\LARGE \bf
	Quaternion Based Camera Pose Estimation \\ From Matched Feature Points
}

\author{Kaveh Fathian, J. Pablo Ramirez-Paredes, Emily A. Doucette, J. Willard Curtis, Nicholas R. Gans
	\thanks{*This work was supported by the OSD sponsored Autonomy Research Pilot Initiative project entitled A Privileged Sensing Framework.}
	\thanks{K. Fathian and N. R. Gans are with the Department of Electrical Engineering, University of Texas at Dallas, Richardson, TX, 75080 USA.~ E-mail: {\tt\small \{kaveh.fathian, ngans\}@utdallas.edu}.        }%
	\thanks{JP Ramirez is with the Research and Engineering Education Facility, University of Florida, Shalimar, FL, 32579, USA.~ E-mail: {\tt\small: jpramirez@ufl.edu}.        }%
	\thanks{E. A. Doucette and J. W. Curtis are with the Air Force Research Laboratory, Munitions Directorate, Eglin AFB, FL, 32542, USA.~ E-mail: {\tt\small \{emily.doucette, jess.curtis\}@us.af.mil}.			}.%
}

\begin{document}

\maketitle
\thispagestyle{empty}
\pagestyle{empty}

\begin{abstract}
We present a novel solution to the camera pose estimation problem, where  rotation and translation of a camera between two views are estimated from matched feature points in  the images. The camera pose estimation problem is traditionally solved via algorithms that are based on the essential matrix or the Euclidean homography. With six or more feature points in general positions in the space, essential matrix based algorithms can recover a unique solution. However, such algorithms fail when points are on critical surfaces (e.g., coplanar points) and homography should be used instead. By formulating the problem in quaternions and decoupling the rotation and translation estimation, our proposed algorithm works for all point configurations. Using both simulated and real world images, we compare the estimation accuracy of our algorithm with some of the most commonly used algorithms. Our method is shown to be more robust to noise and outliers. For the benefit of community, we have made the implementation of our algorithm available online and free\footnote{The Matlab implementation of QuEst is accessible at \mbox{ \color{blue}\burl{https://goo.gl/QH5qhw}}}.
\end{abstract}

\section{Introduction}
\label{sec:introduction}

Many applications in computer vision and robotics require measurements of the rotation and translation (i.e., \textit{pose}) changes of an object as it moves through an environment.  In photogrammetry, for example, by knowing the pose changes of the camera,  3D model of a scene can be constructed from a set of 2D images \cite{Agarwal2009}. In robotics, pose estimated from images can be used for navigation \cite{gans2009,Kneip2013}, or fused with other sensor measurements (e.g., IMU and GPS) to increase the reliability and accuracy \cite{Tick2013}. Camera pose estimation has further applications in simultaneous localization and mapping (SLAM) \cite{Mur-Artal2015}, autonomous vehicles \cite{Schneider2016}, and augmented reality\cite{Marchand2016}.

\begin{figure}
	\begin{center}
		\includegraphics[trim = 30mm 70mm 30mm 60mm, clip, width=0.5\textwidth] {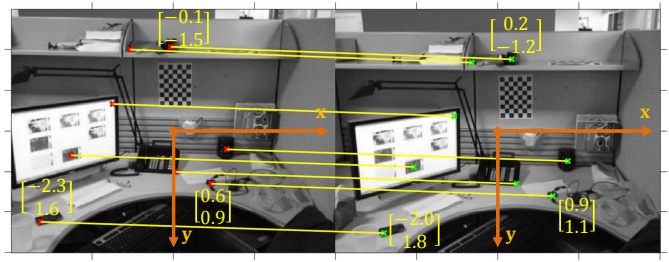}
		\caption{Pairs of matched feature points across two images shown side by side (image courtesy of MathWorks).}
		\label{fig:Features}
	\end{center}
\end{figure}

Camera pose estimation techniques are often based on image features
(e.g., edges, corners, etc., in an image) that can be detected and matched in two or more images \cite{Bay2006, Rublee2011}. {Figure~\ref{fig:Features} } shows an example where feature points are detected and matched as indicated by yellow lines in two images. 
Many existing methods \cite{Higgins81, Demazure1988, Faugeras1993, Malis2000} use the coordinates of feature points on the image to construct the essential/fundamental matrix or the Euclidean homography matrix, from which relative rotation and translation of the camera can be recovered.
Although these approach are fast and easy to implement, they are subject to drawbacks: the essential matrix based algorithms fail when feature points are on critical surfaces (e.g., coplanar points), while homography based algorithms only work when points are coplanar.

In this work, we present a novel formulation of the camera pose estimation problem using quaternions and present a solution to estimate the pose under this formulation. Our approach, which we refer to as the Quaternion Estimation (QuEst) algorithm, does not use the homography or essential matrices, and decouples the estimation of rotation and translation. Consequently, common problems such as degeneracy for special 3D point configurations are avoided. We present two methods to recover the rotation from seven and six matched feature points. We then show how the unique correct solution can be detected from among the set of recovered solutions. 
The performance of QuEst is compared with algorithms that are based on the homography or essential matrix in the presence of noise in image point coordinates. The performance is further vetted by using real world image datasets that come with the ground truth camera pose information.

The main contributions and benefits of the proposed algorithm can be summarized as follows.
\begin{itemize}
	\item As illustrated in Table \ref{tbl:compareAdv}, unlike the homography or essential matrix based algorithms, QuEst provides an accurate pose estimate for both general point configurations and points that are on critical surfaces, e.g., coplanar points. To initialize the bundle adjustment in SLAM applications \cite{Mur-Artal2015}, often heuristic methods are used to detect the coplanarity of the points and select the appropriate  algorithm correspondingly. QuEst can be used to initialize the bundle adjustment regardless of the feature point configuration in the space.
	
	\item By recovering the translation, QuEst simultaneously recovers  depths of the feature points. Therefore, a 3D model of the scene can be reconstructed. The recovered translation and depths share a common scale factor, hence, the magnitude of the recovered translation vector goes to zero as the camera translation between two views approaches zero.  The translation recovered from the essential matrix always has unit norm, which is not desirable in applications such as visual servoing.
	
	\item Tests performed using both simulated and real world images show that the pose recovered from QuEst is more accurate and robust to noise and outliers compared to the existing algorithms. 	
\end{itemize}

\begin{table}
	\caption{QuEst compared with homography and essential matrix based algorithms.}	
	\begin{center}
		\begin{tabular}{c}
			\includegraphics[trim = 40mm 111mm 20mm 40mm, clip, width=0.47\textwidth,keepaspectratio]{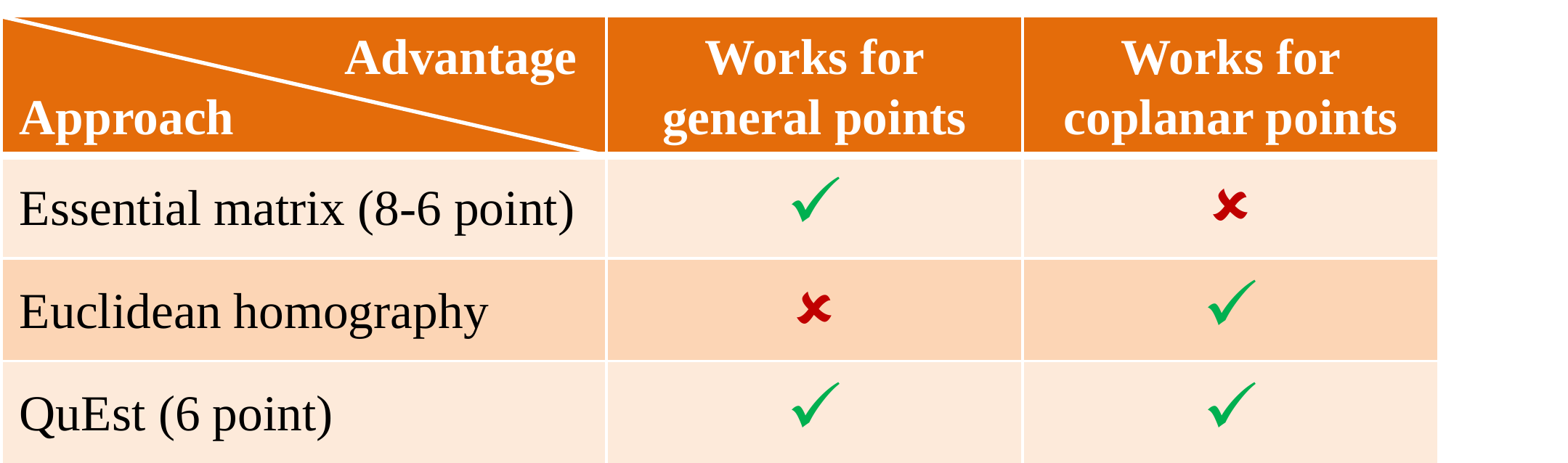}
		\end{tabular}
	\end{center}
\label{tbl:compareAdv}
\end{table}

The rest of the paper is organized as follows.
We briefly review related approaches in Section II, before we
formulate the pose estimation problem using quaternions in Section III. We present the QuEst algorithm in Section IV and evaluate its performance under noise in Section V. In Section IV, we further vet the performance of QuEst using the real world image datasets.

\section{Related Work}
\label{sec:RelatedWork}

Pose estimation using images can be traced back to at least 1913, when Kruppa \cite{kruppa1913} proved that two camera views of five 3D points could be used to estimate the translation and rotation separating two camera views.
However, the lack of sufficient computational resources limited further development until the landmark introduction of the 
8-point algorithm \cite{Higgins81, Demazure1988}.
The 8-point algorithm used the epipolar constraint, embodied in the essential matrix, to solve a set of linear equations that are generated from a set of 8 or more feature points. 

It has been shown that to ensure a finite number of solutions for the pose estimation problem, a minimum of five feature points are required \cite{Faugeras1989, Demazure1988}. However, in general this is not sufficient to recover the pose uniquely. Philip showed that when 6 or more general points are available, the pose estimation problem is linear and produces a unique solution \cite{Philip1996}.

What previous work have in common is the use of the essential matrix in the problem formulation and the assumption that points are in general positions. In the special case where feature points are coplanar, methods based on the homography matrix should be used instead to estimate the pose \cite{Faugeras1993, Malis2000}. However, for a general point configuration that does not contain at least four coplanar points, homography does not return the correct solution. 

We previously used quaternion formulation to solve the pose estimation via an optimization-based method \cite{Fathian2014}. The drawback of this initial work was its reliance on an accurate initial estimate for the pose. The method proposed here is based on formulating an eigenvalue problem, which does not require an initial guess for the solution.

\section{Camera Pose Estimation}
\label{sec:PoseEstiamtion}

We introduce the notation and assumptions, followed by formulating the  pose estimation problem in quaternions. 

\subsection{Notation and Assumptions}
\label{sec:notations}

Throughout the paper, we assume that the camera calibration matrix is known; this matrix can be easily found through the existing camera calibration routines \cite{Bouguet2015}.

Scalars are represented by lower case (e.g., $s$), vectors by lowercase and bold (e.g., $\mathbf{v}$), and matrices by upper case and bold letters (e.g., $\mathbf{M}$).
All vectors are column vectors. 
The Moore-Penrose pseudo inverse of  matrix $\mathbf{M}$ is shown by $\mathbf{M}^\dagger$.
Binomial coefficients are denoted by ${ {n \choose k} := \frac{n!} {k!(n-k)!} }$.
By norm of a vector, we imply the $l^2$-norm. 
Degree 4 \textit{monomials} in variables $w,\, x,\, y,\, z$ are single term polynomials $w^a x^b y^c z^d$, such that $a+b+c+d=4$, for $a, b, c, d \in \{0,1,2,3,4\}$. 

\subsection{Problem Formulation}
\label{sec:Formulation}

Consider images of a scene taken by a camera at two views (e.g., Fig. \ref{fig:Features}).  Let $\mathbf{R} \in \text{SO}(3)$ and ${ \mathbf{t} \in \mathbb{R}^3 }$ respectively represent the relative rotation and translation of the camera frame between the views.
Assume that feature points are detected and matched, and their $x$-$y$ coordinates are read from the images.  (In practice, the coordinates are in pixels, and should be mapped via the camera calibration matrix to Cartesian coordinates on the image plane.)

\begin{figure}
	\begin{center}
		\includegraphics[trim = 40mm 50mm 40mm 55mm, clip, width=0.32\textwidth] {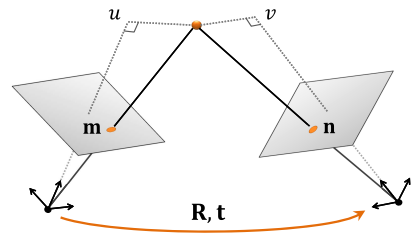}
		\caption{Projection of a 3D point onto the image plane at two views.}
		\label{fig:Projections}
	\end{center}
\end{figure}

For each matched feature point the rigid motion constraint
\begin{equation}
u \, \mathbf{R} \, \mathbf{m} + \mathbf{t}~=~ v \, \mathbf{n}
\label{eq:RigidConstraint} 
\end{equation}
%
must hold, in which $\mathbf{m},\, \mathbf{n} \in \mathbb{R}^3$ are homogeneous coordinates (i.e., a 1 is appended to the $x$-$y$ coordinates) of the feature point  in two images. Scalars  $u$ and $v$ represent depths of the 3D point at each view, and as shown in Fig. \ref{fig:Projections},  are the projections of the point onto the $z$-axis of the camera coordinate frame. Point coordinates $\mathbf{m}$ and $\mathbf{n}$ are known from the images, and the unknowns in \eqref{eq:RigidConstraint} are $u$, $v$, $\mathbf{R}$, and $\mathbf{t}$, which need to be recovered.

We need to point out that in the pose estimation problem, translation and depths of the points can only be recovered up to a scale factor. This can be seen from \eqref{eq:RigidConstraint}, where any constant multiplied into both hand sides can be absorbed by unknown variables $u,\, v$, and $\mathbf{t}$.

\begin{example}
	\label{ex:motionConstraints}
	Consider pictures shown in Fig. \ref{fig:Features}, where feature points are matched and their coordinates on the image are determined. For the matched feature point with coordinates $(-0.1,\, -1.5)$ in the left image and $(0.2,\, -1.2)$ in the right image, from \eqref{eq:RigidConstraint} we get
	\begin{equation}
	u_1 \, \mathbf{R} \, \begin{bsmallmatrix}
	-0.1 \\
	-1.5 \\
	1
	\end{bsmallmatrix} + \mathbf{t}~=~ v_1 \, \begin{bsmallmatrix}
	0.2 \\
	-1.2 \\
	1
	\end{bsmallmatrix},
	\label{eq:pair1} 
	\end{equation}
	%
	where scalar $u_1$ and $v_1$ are the depths of the 3D point at each view. Similarly, for two other matched pairs we can write
	\begin{equation}
	u_2 \, \mathbf{R} \, \begin{bsmallmatrix}
	-2.3 \\
	1.6 \\
	1
	\end{bsmallmatrix} + \mathbf{t}~=~ v_2 \, \begin{bsmallmatrix}
	-2.0 \\
	1.8 \\
	1
	\end{bsmallmatrix},
	\label{eq:pair2} 
	\end{equation}
	%
	\begin{equation}
	u_3 \, \mathbf{R} \, \begin{bsmallmatrix}
	0.6 \\
	0.9 \\
	1
	\end{bsmallmatrix} + \mathbf{t}~=~ v_3 \, \begin{bsmallmatrix}
	0.9 \\
	1.1 \\
	1
	\end{bsmallmatrix},
	\label{eq:pair3} 
	\end{equation}
	%
	%
	where subscripts are used to distinguish the depths of the points (scalars $u$ and $v$). Notice that rotation matrix $\mathbf{R}$ and translation vector $\mathbf{t}$ are the same in all equations. 
\end{example}

Equation \eqref{eq:RigidConstraint} uses the matrix representation of rotation, in which $\mathbf{R}$ is a $3 \times 3$ orthonormal matrix. That is,  ${\mathbf{R}^\top = \mathbf{R}^{-1}}$, and $\det(\mathbf{R}) = 1$. 
Representing the rotation in the matrix form with orthonormality constraints makes the problem very nonlinear and challenging to solve. Instead, we use quaternions, which represent a rotation by four elements $w,\, x,\, y,\, z \in \mathbb{R}$ such that $w^2 + x^2 + y^2 + z^2 = 1$.
Although \eqref{eq:RigidConstraint} can be formulated directly in quaternions, for simplicity and to avoid introducing the quaternion algebra we only mention what is essential to solve the problem here: if $w,\, x,\, y,\, z \in \mathbb{R}$ are elements of a rotation quaternion, the associated rotation matrix is given by
\begin{equation}
\mathbf{R} = \left[\begin{smallmatrix}
w^2+x^2- y^2-z^2 & 2(xy-wz) & 2(xz+wx)\\
2(xy+wz) & w^2-x^2+y^2-z^2 & 2(yz-wz)\\
2(xz-wz) & 2(yz+wy) & w^2-x^2-y^2+z^2
\end{smallmatrix} \right].
\label{eq:quatRotation} 
\end{equation}
%
Quaternions provide a singularity free representation of rotation, and by restricting the first element to nonnegative numbers (i.e., $w \geq 0$), there is a one to one and onto correspondence between rotation matrices and quaternions.

To recover the pose, we first eliminate the unknowns $u,\, v$ and $\mathbf{t}$, and derive a system of equations in terms of the quaternion elements. From solving this system all rotation solution candidates are found. Subsequently, the translation and depths of the points are recovered.
By taking  \eqref{eq:RigidConstraint} for two different feature points and subtracting the equations, $\mathbf{t}$ can be eliminated.
Subsequently, $u$ and $v$ can be eliminated from the resulting equations by noting that they form a null vector for the matrix consisting of the point coordinates and their rotations. The following example illustrates this procedure.

\begin{example}
	\label{ex:removingTranDepth}
	Consider Example \ref{ex:motionConstraints}. By subtracting \eqref{eq:pair2} and \eqref{eq:pair3} from \eqref{eq:pair1}, respectively,  and bringing terms to the left hand side we get
	\begin{equation}
	u_1 \, \mathbf{R}  \left[\begin{smallmatrix}
	-0.1 \\
	-1.5 \\
	1
	\end{smallmatrix}\right] - v_1  \left[\begin{smallmatrix}
	0.2  \\
	-1.2 \\
	1
	\end{smallmatrix}\right] - u_2 \, \mathbf{R}   \left[\begin{smallmatrix}
	-2.3 \\
	1.6  \\
	1
	\end{smallmatrix}\right] + v_2   \left[\begin{smallmatrix}
	-2.0 \\
	1.8  \\
	1
	\end{smallmatrix}\right] = \mathbf{0},
	\label{eq:const1} 
	\end{equation}
	%
	\begin{equation}
	u_1 \, \mathbf{R}  \left[\begin{smallmatrix}
	-0.1 \\
	-1.5 \\
	1
	\end{smallmatrix}\right] - v_1 \left[\begin{smallmatrix}
	0.2  \\
	-1.2 \\
	1
	\end{smallmatrix}\right] - u_3 \, \mathbf{R}  \left[\begin{smallmatrix}
	0.6 \\
	0.9 \\
	1
	\end{smallmatrix}\right] +  v_3 \left[\begin{smallmatrix}
	0.9 \\
	1.1 \\
	1
	\end{smallmatrix}\right] = \mathbf{0},
	\label{eq:const2} 
	\end{equation}
	%
	in which the unknown translation $\mathbf{t}$ has been eliminated. We can represent \eqref{eq:const1} and \eqref{eq:const2} in the matrix-vector form
	\begin{equation}
	\scriptstyle 
	\underbrace{
	\left[\begin{smallmatrix}
	\mathbf{R}  \left[\begin{smallmatrix}
	-0.1 \\
	-1.5 \\
	1
	\end{smallmatrix}\right] & 
	 \left[\begin{smallmatrix}
	-0.2  \\
	1.2   \\
	1
	\end{smallmatrix}\right] &
	\mathbf{R}   \left[\begin{smallmatrix}
	2.3   \\
	-1.6  \\
	1
	\end{smallmatrix}\right] &
	\left[\begin{smallmatrix}
	-2.0 \\
	1.8  \\
	1
	\end{smallmatrix}\right] &
	\mathbf{0} & 
	\mathbf{0}			 \\ \\ 
	\mathbf{R}  \left[\begin{smallmatrix}
	-2.0 \\
	1.8  \\
	1
	\end{smallmatrix}\right] & 
	\left[\begin{smallmatrix}
	-0.2  \\
	1.2   \\
	1
	\end{smallmatrix}\right] &
	\mathbf{0} &
	\mathbf{0} &
	\mathbf{R} \left[\begin{smallmatrix}
	-0.6 \\
	-0.9 \\
	1
	\end{smallmatrix}\right] &
	\left[\begin{smallmatrix}
	0.9 \\
	1.1 \\
	1
	\end{smallmatrix}\right]
	\end{smallmatrix}\right] 
	}_{\mathbf{M}}
	\left[\begin{smallmatrix}
	\\
	u_1 \\ \\
	v_1 \\ \\
	u_2 \\ \\
	v_2 \\ \\
	u_3 \\ \\
	v_3
	\end{smallmatrix}\right] = \,\mathbf{0},
	\label{eq:ElimDepth} 
	\end{equation}
	%
	%
	which implies that matrix $\mathbf{M} \in \mathbb{R}^{6\times 6}$ has a null vector. Therefore, its determinant must be zero. Calculating determinant of $\mathbf{M}$ with $\mathbf{R}$ given in the parametric form \eqref{eq:quatRotation} gives
	\begin{align}
	-0.1 \, w^4 + 2.4 \, w^3 x + 35.6 \, w^2 x^2  -11.4 \, w x^3 + 0.3 \, x^4 +  \dots \nonumber \\ 
	-19.8 \, y^2 z^2  -168 \, w z^3 + 9.4 \, x z^3  -17.8 \, y z^3 + 0.3 \, z^4 = 0. &
	\label{eq:QuarticConstr} 
	\end{align}
	%
	Equation \eqref{eq:QuarticConstr} consists of all degree 4 monomials in $w,\, x,\, y,\, z$ (i.e., terms such as $w^4, w^3 x, w^2 x^2, \dots$), with coefficients that depend on the feature point coordinates. Note that since $\mathbf{M}$ is a $6 \times 6$ matrix, one may expect its determinant to have degree six monomials, however, due to special structure of $\mathbf{M}$, it is always possible to factor out $w^2 + x^2 + y^2 + z^2$ from the determinant expression. Since $w^2 + x^2 + y^2 + z^2 = 1$, the degree four polynomial equation \eqref{eq:QuarticConstr} follows. 
\end{example}

What we showed in Example \ref{ex:removingTranDepth} was that three matched feature points generate a polynomial equation of the form  \eqref{eq:QuarticConstr}. This equation is in terms of degree four monomials in  $w,\, x,\, y,\, z$. Note that there are 35 such monomials, and their coefficients are in terms of the feature point coordinates. In practice, we do not need to calculate the determinant of $\mathbf{M}$ to find these coefficients. By replacing the feature point coordinates with symbolic expressions the determinant can be computed symbolically, and explicit formulas for the coefficients can be derived. By substituting the numerical values of point coordinates in these formulas the coefficients are calculated directly. Due to the space limitation we do not give the explicit formulas here.

Since any three feature points give a polynomial equation of the form \eqref{eq:QuarticConstr}, from $n$ points ${n \choose 3}$ equations can be generated. These equations can be stacked into a matrix-vector form, where the vector consists of the unknown monomial terms. The following example illustrates this point.

\begin{example}
	\label{ex:MatrixRepresentation}
	Consider ${6 \choose 3} = 20$ polynomial equations of the form \eqref{eq:QuarticConstr}, generated from 6 feature points.  These polynomial equations can be represented in the matrix-vector form
	\begin{equation}
	\underbrace{
	\begin{bmatrix}
	-0.1   &  2.4   & 35.6   & \cdots  & 0.3   \\
	-0.1   &  0.1   & -7.6   & \cdots  & -0.2   \\
	\vdots & \vdots & \vdots & \ddots  & \vdots \\
	0.0   &  -0.2   & 4.6   & \cdots  & 0.1    \\
	\end{bmatrix}
	}_{\mathbf{A}}
	\underbrace{
	\begin{bmatrix}	
	w^4      \\
	w^3 x    \\
	w^2 x^2  \\
	\vdots   \\
	z^4
	\end{bmatrix}
	}_{\mathbf{x}}
	 = \mathbf{0}
	\label{eq:polySystem} 
	\end{equation}
	%
	where the coefficient matrix $\mathbf{A} \in \mathbb{R}^{20\times 35}$ depends on the feature point coordinates, and vector $\mathbf{x} \in \mathbb{R}^{35}$ consists of all degree 4 monomials. Our goal is to find all $w,\, x,\, y,\, z$, for which \eqref{eq:polySystem} is satisfied.
\end{example}

The problem of recovering the rotation is henceforth equivalent to solving
a system of equations of the form \eqref{eq:polySystem}, where the goal is to find all $\mathbf{x}$ for which
\begin{equation}
\mathbf{A} \, \mathbf{x} = \mathbf{0}
\label{eq:Ax} 
\end{equation}
%
is satisfied. In \eqref{eq:Ax}, the coefficient matrix $\mathbf{A}$ is known from the feature point coordinates, and vector $\mathbf{x}$ is unknown with entries in degree four monomials of $w,\, x,\, y,\, z$.

\section{The QuEst Algorithm}
\label{sec:QuEst}

In what follows we first show how rotation solution candidates can be recovered from 7 and 6 matched feature points. Given a rotation solution candidate, it is then shown how the associated translation vector and depths are recovered. Lastly, we show how the unique solution can be distinguished by discarding the physically infeasible solution candidates.
The Matlab implementation of QuEst is accessible at \mbox{ \color{blue}\burl{https://goo.gl/QH5qhw}}.

We will not discuss why the pose estimation problem has always more than one \textit{mathematically} feasible solution (e.g., 2 for general points and 4 for coplanar points) since these results are well-known. Interested readers are referred to \cite{Ma2003} for further discussion and mathematical proofs on the number of solutions.

\subsection{Recovering Rotation From 7 Points}
\label{subsec:7point}

Consider the system of equations \eqref{eq:Ax} for 7 matched feature points. Since 7 points generate ${7 \choose 3} = 35$ equations, in this case $\mathbf{A}$ is a $35\times 35$ matrix. Due to the mathematical multiplicity of solutions however, $\mathbf{A}$ cannot be full rank (otherwise, only one solution exists, which is a contradiction).

Let us arrange the entries of $\mathbf{x} \in \mathbb{R}^{35}$ in \eqref{eq:Ax} such that ${\mathbf{x} = 
\begin{bmatrix}
\mathbf{x}_1 \\
\mathbf{x}_2
\end{bmatrix} }$, where $\mathbf{x}_1 \in \mathbb{R}^{4}$ and $\mathbf{x}_2 \in \mathbb{R}^{31}$ are defined as
\begin{equation}
\mathbf{x}_1 := 
\begin{bmatrix}
w^4 \\
w^3 x \\
w^3 y \\
w^3 z  
\end{bmatrix}, \quad
\mathbf{x}_2 := 
\begin{bmatrix}
x^3 w \\
x^4 \\
x^3 y \\
x^3 z \\
\vdots
\end{bmatrix}.
\label{eq:7ptx} 
\end{equation}
Let $\mathbf{A} = [\mathbf{A}_1 ~ \mathbf{A}_2]$, where $\mathbf{A}_1 \in \mathbb{R}^{35\times 4}$ is the first 4 columns of $\mathbf{A}$, and $\mathbf{A}_2 \in \mathbb{R}^{35\times 31}$ is the remaining part. Equation \eqref{eq:Ax} is equivalent to
\begin{equation}
\mathbf{A}_1 \, \mathbf{x}_1 + \mathbf{A}_2 \, \mathbf{x}_2
= \mathbf{0},
\label{eq:7ptDecomp} 
\end{equation}
%
from which, by multiplying the pseudo inverse of $\mathbf{A}_2$ from the left, we obtain\footnote{For a general point configuration $\mathbf{A}_2$ has rank 31, and thus $\mathbf{A}_2^\dagger \mathbf{A}_2 = \mathbf{I}$, where $\mathbf{I}$ is the identity matrix.}
\begin{equation}
\mathbf{x}_2  = -\mathbf{A}_2^\dagger \, \mathbf{A}_1 \, \mathbf{x}_1.   
%
\label{eq:7ptA1A2} 
\end{equation}
%
Let $\mathbf{\bar{B}} := -\mathbf{A}_2^\dagger \, \mathbf{A}_1 \in \mathbb{R}^{31\times 4}$. The first 4 rows of \eqref{eq:7ptA1A2} imply
\begin{equation}
\begin{bmatrix}
x^3 w \\
x^4 \\
x^3 y \\
x^3 z  
\end{bmatrix}  = 
\mathbf{B}
\begin{bmatrix}
w^4 \\
w^3 x \\
w^3 y \\
w^3 z 
\end{bmatrix},
\label{eq:7ptB} 
\end{equation}
%
where $\mathbf{B} \in \mathbb{R}^{4\times 4}$ is the matrix consisting of the first 4 rows of $\mathbf{\bar{B}}$.
By factoring $x^3$ from the left hand side vector and $w^3$ from the right hand side vector of \eqref{eq:7ptB} we get
\begin{equation}
\frac{x^3}{w^3}
\begin{bmatrix}
w \\
x \\
y \\
z  
\end{bmatrix}  = 
\mathbf{B}
\begin{bmatrix}
w \\
x \\
y \\
z 
\end{bmatrix},
\label{eq:7ptEig} 
\end{equation}
%
which is an eigenvalue problem of the form $\lambda \mathbf{v} = \mathbf{B} \mathbf{v}$, with $\lambda = \frac{x^3}{w^3}$ and $\mathbf{v} = [w~ x~ y~ z]^\top$. Hence, 4 solution candidates are found by calculating the (unit norm) eigenvectors of $\mathbf{B}$.

We should mention that the choice of $\mathbf{x}_1$ and $\mathbf{x}_2$ are somewhat arbitrary. For example, we could have chosen $\mathbf{x}_2$ as $\mathbf{x}_2 := \begin{bmatrix}
y^3 w & y^3 x & y^4 & y^3 z & \dots
\end{bmatrix}^\top$, and derive a similar eigenvalue problem with $\lambda = \frac{y^3}{w^3}$. We will later use this fact to distinguish the unique solution.

\subsection{Recovering Rotation From 6 Points}
\label{subsec:6point}

Consider equation \eqref{eq:Ax} for 6 feature points. Since 6 feature points generate ${6 \choose 3} = 20$ equations, in this case $\mathbf{A}$  is a $20\times 35$ full rank matrix.

We split $\mathbf{x} \in \mathbb{R}^{35}$ into two vectors $\mathbf{x}_1 \in \mathbb{R}^{20}$ and  $\mathbf{x}_2 \in \mathbb{R}^{15}$, where ${\mathbf{x} = 
	\begin{bmatrix}
	\mathbf{x}_1 \\
	\mathbf{x}_2
\end{bmatrix}}$, $\mathbf{x}_1$ is the vector of all monomials that contain a power of $w$ (e.g., $w^4,\, w^3 x,\, w^3 y,\, \dots,\, w y^3,\, w z^3$), and  $\mathbf{x}_2$ consists of the rest of the monomials (e.g., $x^4,\, x^3 y,\, x^2 y^2,\, \dots,\, y z^3, z^4$).
Let ${\mathbf{A} = [\mathbf{A}_1 ~ \mathbf{A}_2]}$, where ${\mathbf{A}_1 \in \mathbb{R}^{20\times 20} }$ consists of the first 20 columns of $\mathbf{A}$, and $\mathbf{A}_2 \in \mathbb{R}^{20\times 15}$ is the remaining part. Equation \eqref{eq:Ax} is equivalent to
\begin{equation}
\mathbf{A}_1 \, \mathbf{x}_1 + \mathbf{A}_2 \, \mathbf{x}_2 = \mathbf{0},
\label{eq:6ptDecomp} 
\end{equation}
%
from which, by multiplying the pseudo inverse of $\mathbf{A}_2$, we obtain
\begin{equation}
\mathbf{x}_2  = -\mathbf{A}_2^\dagger \mathbf{A}_1 \, \mathbf{x}_1.
\label{eq:6ptA1A2} 
\end{equation}
%
Since $\mathbf{x}_1$ consists of monomials that have at least one power of $w$, we can factor out $w$ and represent the remaining vector by $\mathbf{v}  \in \mathbb{R}^{20}$, i.e.,  $\mathbf{v} = \frac{1}{w} \mathbf{x}_1$. Thus, \eqref{eq:6ptA1A2} can be written as
\begin{equation}
\mathbf{x}_2  = w \, \mathbf{\bar{B}}  \, \mathbf{v},
\label{eq:6ptB} 
\end{equation}
%
where $\mathbf{\bar{B}} := -\mathbf{A}_2^\dagger \mathbf{A}_1 \in \mathbb{R}^{15\times 20}$.

Equation \eqref{eq:6ptB} allows us to construct an eigenvalue problem of the form $\lambda \mathbf{v} = \mathbf{B} \mathbf{v}$, with $\mathbf{B} \in \mathbb{R}^{20\times 20}$. Indeed, let us choose  $\lambda = \frac{x}{w}$, and consider the eigenvalue problem
\begin{equation}
x \, \mathbf{v}  = w \, \mathbf{B} \, \mathbf{v}.
\label{eq:6ptEig} 
\end{equation}
%
The entries of vector $x \, \mathbf{v}$ either belong to $\mathbf{x}_2$ or $\mathbf{x}_1$. For entries that belong to $\mathbf{x}_2$, the associated rows of $\mathbf{B}$ are chosen from the corresponding rows of $\mathbf{\bar{B}}$ in \eqref{eq:6ptB}. For entries that belong to $\mathbf{x}_1$, rows of $\mathbf{B}$ are chosen as $[0\, \dots\, 0~ 1~ 0\, \dots\, 0]$. The following example illustrates this procedure.

\begin{example}
	\label{ex:6ptExample}
	Suppose entries of $\mathbf{x}_1$ and $\mathbf{x}_2$ are arranged as
	\begin{equation}
	\mathbf{x}_1 = \begin{bmatrix}	
	w^3 x      \\
	w^2 x^2    \\
	w x^3    \\
	\scriptstyle \vdots   \\
	w z^3     
	\end{bmatrix}, \quad
	\mathbf{x}_2 = \begin{bmatrix}	
	z^4      \\
	x z^3    \\
	y z^3    \\
	\scriptstyle \vdots   \\
	x^4
	\end{bmatrix}, \quad
	\mathbf{v} = \frac{1}{w} \mathbf{x}_1 =  \begin{bmatrix}	
	w^2 x      \\
	w x^2    \\
	x^3    \\
	\scriptstyle \vdots   \\
	z^3 
	\end{bmatrix},
	\label{eq:6ptExVecs} 
	\end{equation}
	%
	and assume that from the feature point coordinates we have derived \eqref{eq:6ptB} as
	\begin{equation}
	\mathbf{x}_2 = 
	\begin{bmatrix}	
	z^4      \\
	x z^3    \\
	y z^3    \\
	\scriptstyle \vdots   \\
	x^4
	\end{bmatrix} = 
	w\, \begin{bmatrix}	
	-0.1  &  5.2  & 22.9 & \scriptstyle \cdots  &  14    \\
	0.1  &  -6.7  & -4.4 & \scriptstyle \cdots  &  -15    \\
	0.0  &  -4.7  & -1.1 & \scriptstyle \cdots  &  -11    \\
	\scriptstyle \vdots &\scriptstyle \vdots &\scriptstyle \vdots & \scriptstyle \ddots & \scriptstyle \vdots  \\
	-0.1  &  -0.4  & 26.2 & \scriptstyle \cdots &  46
	\end{bmatrix} \mathbf{v}.
	\label{eq:6ptx2} 
	\end{equation}
	%
	From \eqref{eq:6ptx2} we can construct the eigenvalue problem  \eqref{eq:6ptEig} as
	\begin{equation}
	x\, \mathbf{v} = 
	\begin{bmatrix}	
	w^2 x^2      \\
	w x^3    \\
	x^4    \\
	\scriptstyle \vdots   \\
	x z^3 
	\end{bmatrix} = 
	w\, \begin{bmatrix}
	0      &  1     & 0      & \scriptstyle \cdots  &  0     \\
	0      &  0     & 1      & \scriptstyle \cdots  &  0     \\
	-0.1   &  -0.4  & 26.2   & \scriptstyle \cdots  &  46	\\
	\scriptstyle \vdots &\scriptstyle \vdots &\scriptstyle \vdots & \scriptstyle \ddots & \scriptstyle \vdots   \\
	0.1    &  -6.7  & -4.4   & \scriptstyle \cdots  & -15    \\
	\end{bmatrix} \mathbf{v},
	\label{eq:} 
	\end{equation}
	%
	where the first two entries of $x \, \mathbf{v}$ belong to $\mathbf{x}_1 = w \, \mathbf{v}$, and hence their associated rows in $\mathbf{B}$ consist of zeros except for a single one entry.  The third and last entries of $x \, \mathbf{v}$ belong to $\mathbf{x}_2$, and their associated rows come from $\mathbf{\bar{B}}$ in \eqref{eq:6ptx2}.
\end{example}

Once the eigenvalue problem \eqref{eq:6ptEig} is constructed, 20 solution candidates for $\mathbf{v}$ are derived by computing the eigenvectors of $\mathbf{B}$. For each solution candidate, $w,\, x,\, y,\, z$ are found by calculating the third root of the $w^3,\, x^3,\, y^3,\, z^3$ entries in $\mathbf{v}$. 
The recovered solution can be normalized to meet the unit norm constraint $w^2 + x^2 + y^2 + z^2 = 1$. 
Notice that by choosing $\mathbf{x}_1$ or $\lambda$ differently (e.g., $\lambda = \frac{y}{w}$) it is possible to derive different eigenvalue problems of the form \eqref{eq:6ptB}.

\subsection{Recovering Translation and Depths}
\label{sec:tran}

Once quaternion elements $w,\, x,\, y,\, z$ are recovered, the corresponding rotation matrix $\mathbf{R}$ is given by \eqref{eq:quatRotation}. Having $\mathbf{R}$, the rigid motion constraint $u \, \mathbf{R} \, \mathbf{m} + \mathbf{t} = v \, \mathbf{n}$ can now be written for all matched feature points, and stacked into the matrix-vector form
\begin{equation}
\underbrace{
\begin{bsmallmatrix}
	\mathbf{I} & \mathbf{R} \mathbf{m}_1 & -\mathbf{n}_1 & \mathbf{0} & \mathbf{0} & \cdots & \mathbf{0} & \mathbf{0} \\
	\mathbf{I} & \mathbf{0} & \mathbf{0} & \mathbf{R} \mathbf{m}_2 & -\mathbf{n}_2 &  & \mathbf{0} & \mathbf{0} \\
	\vdots  &   & \vdots  &   &   & \ddots &    &  \vdots  \\
	\mathbf{I} & \mathbf{0} & \mathbf{0} & \mathbf{0} & \mathbf{0} & \cdots & \mathbf{R} \mathbf{m}_k & -\mathbf{n}_k \\
\end{bsmallmatrix} 
}_{\mathbf{C}}
\underbrace{
\begin{bsmallmatrix}
\mathbf{t} \\
u_1 \\
v_1 \\
u_2 \\
v_2 \\
\vdots \\
u_k \\
v_k \\
\end{bsmallmatrix}
}_{\mathbf{y}}
 = \mathbf{0}
\label{eq:trans}
\end{equation}
%
where $\mathbf{I} \in \mathbb{R}^{3\times 3}$ is the identity matrix,  $k$ is the number feature points, $\mathbf{C} \in \mathbb{R}^{3k \,\times\, 2k+3}$, and $\mathbf{y} \in \mathbb{R}^{2k+3}$. Equation \eqref{eq:trans} implies that $\mathbf{y}$ is in the null space of $\mathbf{C}$. Thus, $\mathbf{y}$ can be found by calculating the rightmost singular vector of $\mathbf{C}$ (i.e., eigenvector of $\mathbf{C}^\top \mathbf{C}$ corresponding to the zero eigenvalue). Notice that $\mathbf{y}$ consists of the translation vector and feature point depths. Therefore, these parameters are recovered simultaneously and with a common scale factor.

\subsection{The Unique Solution}
\label{sec:UniqeSol}

Before we proceed with finding the unique solution, we need to briefly talk about the \textit{critical surfaces}. Critical surfaces are special configurations of 3D points in the space for which one cannot distinguish a unique solution. (In this case the problem always has more than one physically realizable solution.) 
Perhaps the most important and practical example of such surfaces is when all 3D points lie on a plane, i.e., \textit{coplanar} points. Coplanar points are  abundant in aerial images (due to large distance of points from the camera) or images of the man-made environments (due to points lying on walls, floor, etc.). In what follows we will discuss the case of general and coplanar points separately.

Since feature points that are reflected with respect to the origin of the camera frame produce the same image, to detect the physically infeasible solutions one should check the \textit{chirality}. 
Solution with the wrong chirality correspond to points that are behind the camera, and therefore have negative depths. Hence, physically infeasible solution candidates can be detected and discarded after recovering the depths.

\subsubsection{General points}

As discussed at the end of Sections \ref{subsec:7point} and \ref{subsec:6point}, by choosing other values for $\lambda$ (e.g. ${ \lambda = \frac{y}{w},\, \frac{z}{w} }$) similar eigenvalue problems of the form \eqref{eq:7ptEig} and \eqref{eq:6ptEig} can be derived. Since the correct solution must satisfy the eigenvalue problem regardless of the chosen $\lambda$, solution candidates that do not satisfy $\lambda \, v = \mathbf{B} \, v$ for all values of $\lambda$ can be discarded. In this case, two mathematically feasible solutions remain, from which the unique solution is determined by checking the chirality.

\subsubsection{Coplanar points}
   
When points are coplanar (or more generally lie on critical surfaces), four mathematically feasible solutions remain after checking  $\lambda \, v = \mathbf{B} \, v$ for different values of $\lambda$. Two of these solutions can be discarded by checking the chirality. To determine the solution uniquely further information is required (e.g., a third view or the normal vector to the plane). 

Although in theory the methods discussed above can eliminate infeasible solution candidates, in practice pixelization noise and matching imperfections may result in choosing the wrong solution. For instance, in applications where \textit{parallax} is small (i.e., translation between the views is much smaller than the average distance of the 3D points to the camera), recovering the depths becomes an ill-conditioned problem. Thus, checking the chirality may result in the wrong conclusion. Furthermore, with noise, coplanar points may appear as if they are at general positions and instead of two only one solution is returned. It is therefore recommended to always keep the best four solution candidates, and use the Random Sample Consensus (RANSAC) algorithm to find the unique solution.

\section{Noise and Time Benchmarks}
\label{sec:}

We benchmark our proposed algorithm against some of the most well-known algorithms such as the essential matrix 8-point \cite{Higgins81}, 7-point, and 6-point algorithms \cite{Ma2003}, and the Euclidean homography algorithm \cite{Hartley2004}. For the first three algorithms the Stewenius's implementation \cite{n-pt-implementation}, and for the latter Hartley's implementation in Matlab are used.  
We refer to the algorithms presented in this paper for 6 and 7 points respectively as QuEst 6 and QuEst 7.

Each Monte Carlo simulation consists of eight randomly generated 3D points with uniform distribution inside a rectangular parallelepiped in front of the camera at the initial location.  The camera is moved to a second location by a random translation and rotation quaternion, with uniform distribution within a bounded box of $\mathbb{R}^3$ and on the 3-sphere, respectively. Coordinates of the feature points on the image plane are computed by projecting the 3D points on the image planes. Each algorithm is provided with the minimum number of points it requires to compute the pose.

\subsection{Noise Benchmarks}
\label{subsec:NoiseBenchmark}

To evaluate the performance under noise, Gaussian noise with zero mean and standard deviation ranging from 0 to 10 pixels is added to all image coordinates. The noise standard deviation is increased by 0.1 pixel increments, and for each noise increment 100 simulations are generated.
As mentioned in Section \ref{sec:UniqeSol}, the chirality condition is sensitive to noise, so to avoid choosing the wrong solution, the solution candidate that is closest to the ground truth is chosen as the best  pose estimate for each algorithm. The estimation error for rotation is defined by
\begin{equation}
\rho(\mathbf{q},\mathbf{q}^*) \,=\, \frac{1}{\pi} \arccos(\text{dot}(\mathbf{q}, \, \mathbf{q}^*)) \in [0,1]
\label{eq:RotMetric}
\end{equation}
where $\mathbf{q}= [w~ x~ y~ z]^\top$ is the rotation estimated from the noisy images, and  $\mathbf{q^*}$ is the ground truth rotation in quaternions. Note that \eqref{eq:RotMetric} defines a metric on the rotation quaternion space \cite{Huynh2009}. Similarly, the estimation error for translation is defined by
\begin{equation}
\rho(\mathbf{t_n},\mathbf{t^*_n}) \,=\, \frac{1}{\pi} \arccos(\text{dot}(\mathbf{t_n}, \, \mathbf{t^*_n})) \in [0, 1]
\label{eq:TranMetric}
\end{equation}
where  $\mathbf{t_n}$ and $\mathbf{t^*_n}$ are the estimated translation vector and the ground truth, respectively, normalized to have unit norm  (because the magnitude of the recovered translation vector can vary depending on the algorithm).

\begin{figure}
	\begin{center}
		\includegraphics[trim = 0mm 0mm 0mm 0mm, clip, width=0.48\textwidth, keepaspectratio] {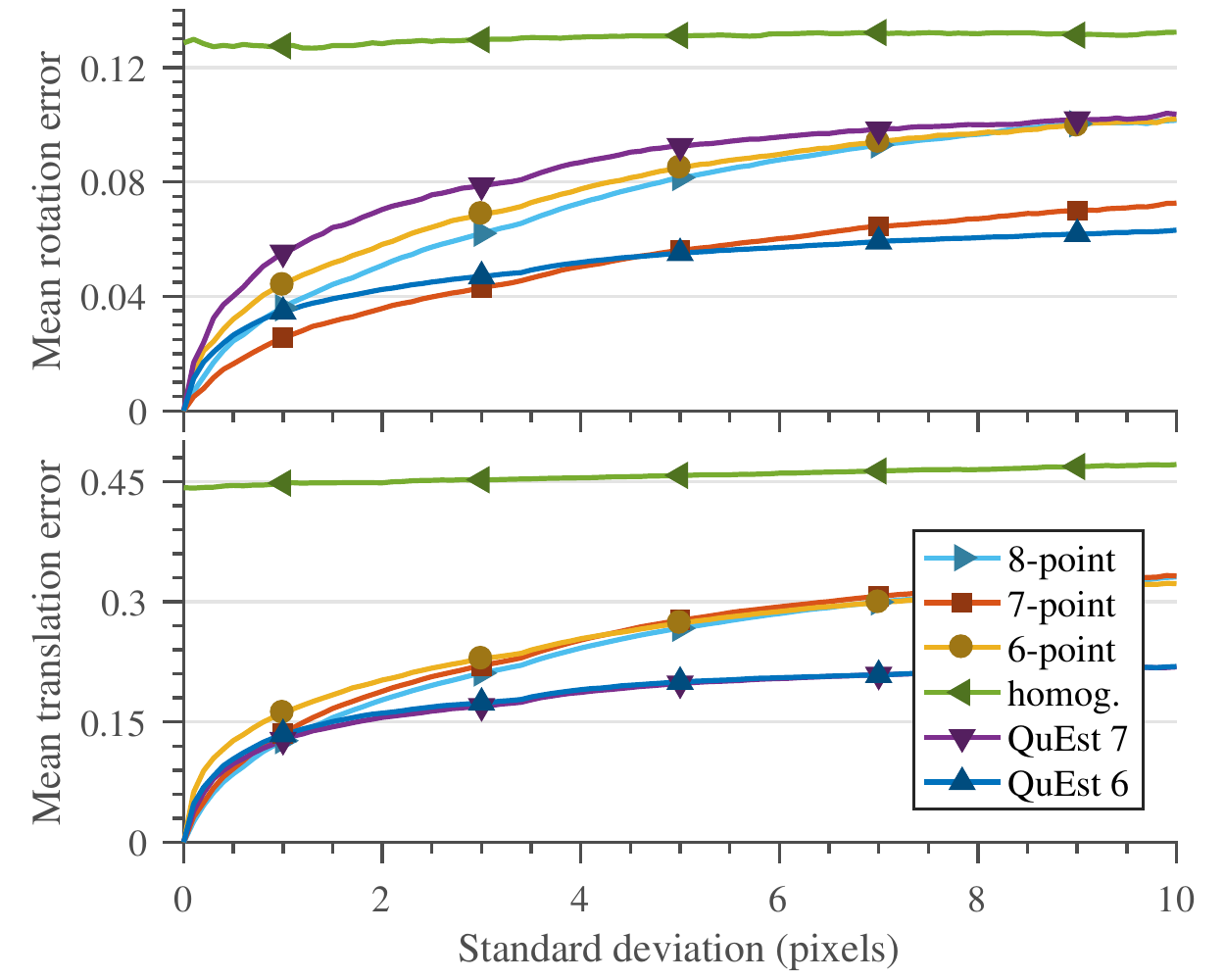} 
		\caption{Comparison under Gaussian noise on feature point coordinates when points are in general 3D configuration.}
		\label{fig:Noise3D}
	\end{center}
\end{figure}

Figure \ref{fig:Noise3D} shows the mean rotation and translation estimation errors at different noise standard deviations for all algorithms.
Since the feature points are randomly generated and are generally non-coplanar, the homography algorithm, which only works for coplanar points, fails to correctly estimate the pose. Essential matrix based algorithms however are not affected. QuESt 6 and 7-point algorithms have the best performance for rotation, while QuESt 6 and QuESt 7 have the best performance for translation estimates. Unlike the 7-point algorithm, translation estimated by QuESt 7 benefits from the extra points and is comparable to QuEst 6. 
We should mention that for large noise standard deviations (e.g., 10 pixels) the results can be interpreted as how robust an algorithm is to incorrectly matched feature points.

\begin{figure}
	\begin{center}
		\includegraphics[trim = 0mm 0mm 0mm 0mm, clip, width=0.48\textwidth, keepaspectratio] {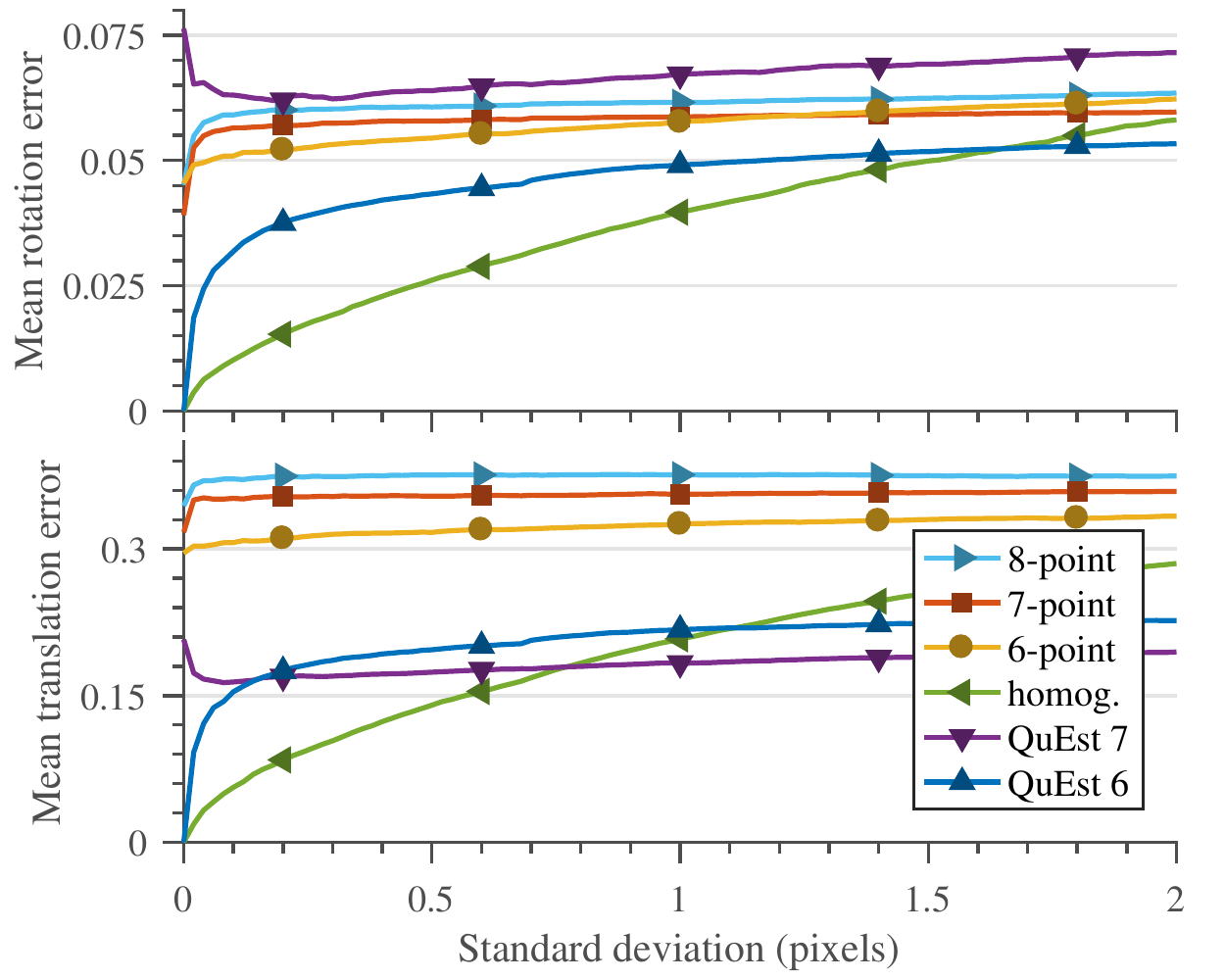} 
		\caption{Comparison under Gaussian noise on feature point coordinates when points are in coplanar configuration.}
		\label{fig:Noise2D}
	\end{center}
\end{figure}

To analyze the performance when points are on critical surfaces, the previous analysis is repeated for coplanar points, where the points are chosen randomly on a bounded plane with uniform distribution. The mean of the rotation and translation estimation errors for noise standard deviation varying from 0 to 2 is shown in Fig. \ref{fig:Noise2D}. As can be seen from the figure, homography shows the best noise resilience when the standard deviation is small (approximately 1 to 1.5 pixels). This is because the homography algorithm is specifically designed to recover the pose when points are coplanar. QuEst 6 has the next best estimation accuracy. 
When points are on critical surfaces matrix $\mathbf{A}_2$ in \eqref{eq:7ptDecomp} loses rank and becomes rank 27. Hence, multiplication by  $\mathbf{A}_2^\dagger$ will not result in \eqref{eq:7ptDecomp}, and QuEst 7 fails to recover the pose. On the other hand, $\mathbf{A}_2$ used for QuEst 6 in \eqref{eq:6ptDecomp} remains full rank due to having smaller dimensions. Lastly, none of the algorithms that are based on the essential matrix can recover the pose in this case, regardless of the number of points used in the algorithm or the magnitude of noise (see \cite{Ma2003} for further explanation).

In conclusion, QuEst 6 shows the best performance since the pose is estimated correctly regardless of the 3D point configuration, and the estimation is robust to noise and outliers.

\subsection{Time Benchmarks}
\label{subsec:TimeBenchmark}

Table \ref{tbl:TimeComp} lists the average execution time of all algorithms in milliseconds, where 1000 Monte Carlo simulations with both coplanar and general points and various noise magnitudes are used to generate the results. All algorithms are implemented as Mex files in Matlab, and tested on the same platform with Intel's 4th Gen i-7 CPU. Algorithms that use homography or essential matrix have smaller execution time since fewer operations are needed to estimate these matrices. QuEst has a larger execution time since the rotation and translation are recovered independently. We should emphasize that although QuEst is not at fast as other algorithms, it is fast enough to be used with RANSAC in real-time applications. Furthermore, since QuEst recovers the pose regardless of the 3D point configuration, no prior effort is required to detect the coplanarity and choose the appropriate algorithm correspondingly.

\begin{table}
	\caption{Average execution time of essential matrix based algorithms, homography algorithm, and QuEst algorithms.}
	\begin{center}
		\begin{tabular}{c}
			\includegraphics[trim = 45mm 85mm 45mm 40mm, clip, width=0.4\textwidth,keepaspectratio]{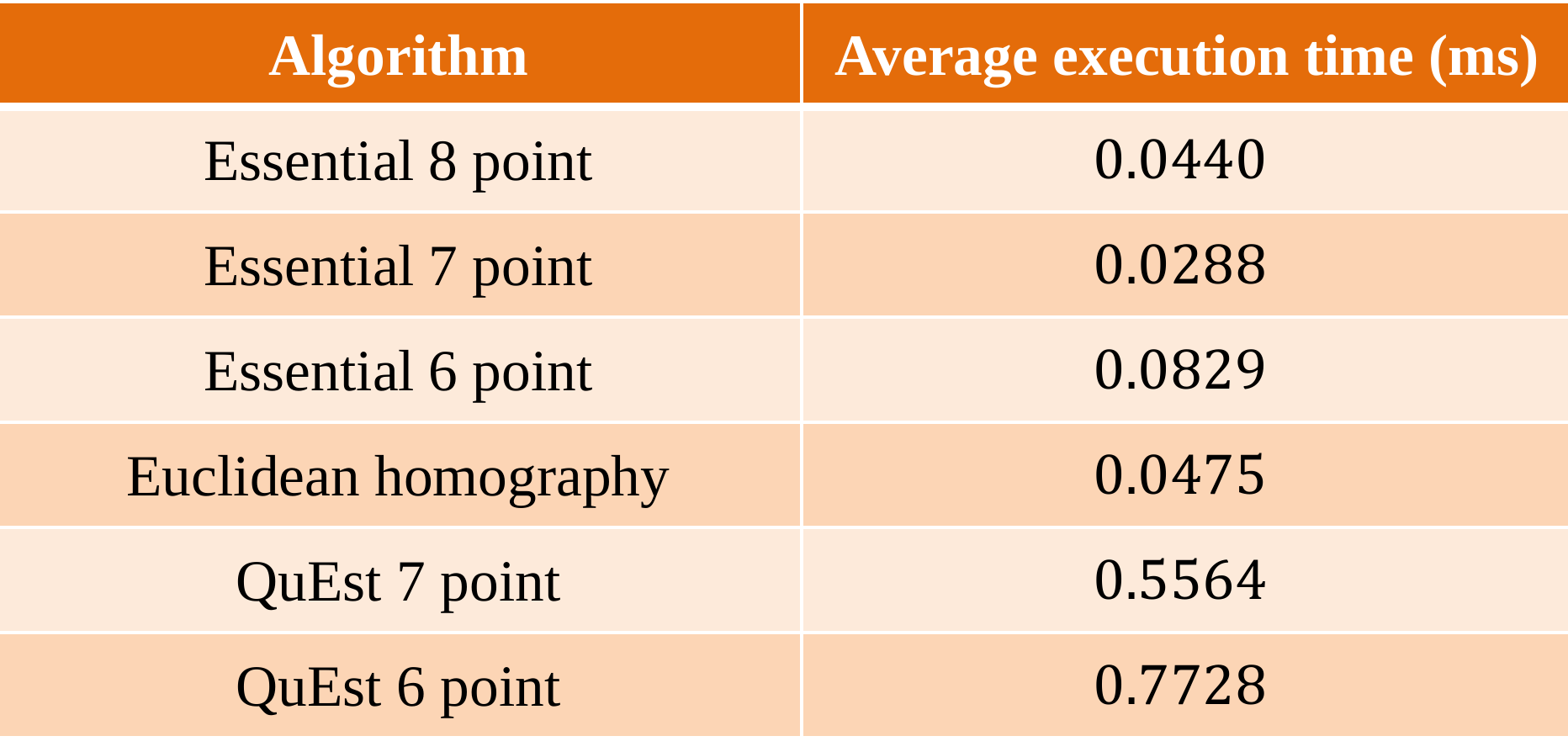}
		\end{tabular}
		\label{tbl:TimeComp}
	\end{center}
\end{table}

\section{Real World Performance}
\label{sec:}

To further assess the accuracy and robustness of QuEst, images from four real world datasets are used to estimate the pose and compare the results with the ground truth that is provided by the dataset. Datasets used for the comparison are ICL \cite{Handa2014}, KITTI \cite{Geiger2012}, NAIST \cite{Tamura2009}, and TUM \cite{Sturm2012}, where SURF image feature points are extracted and matched between two consecutive keyframes. 
Each algorithm is provided with the minimum number of points it requires to estimate the pose, where points with the highest matching score are chosen.  To test the robustness of the algorithms, RANSAC is not used, and the provided set of matched points can occasionally have outliers.

\begin{figure*}[]
	\begin{center}
		\begin{subfigure}[b]{1\textwidth}
			\includegraphics[trim = 0mm 0mm 10mm 0mm, clip, width=1\textwidth, keepaspectratio] {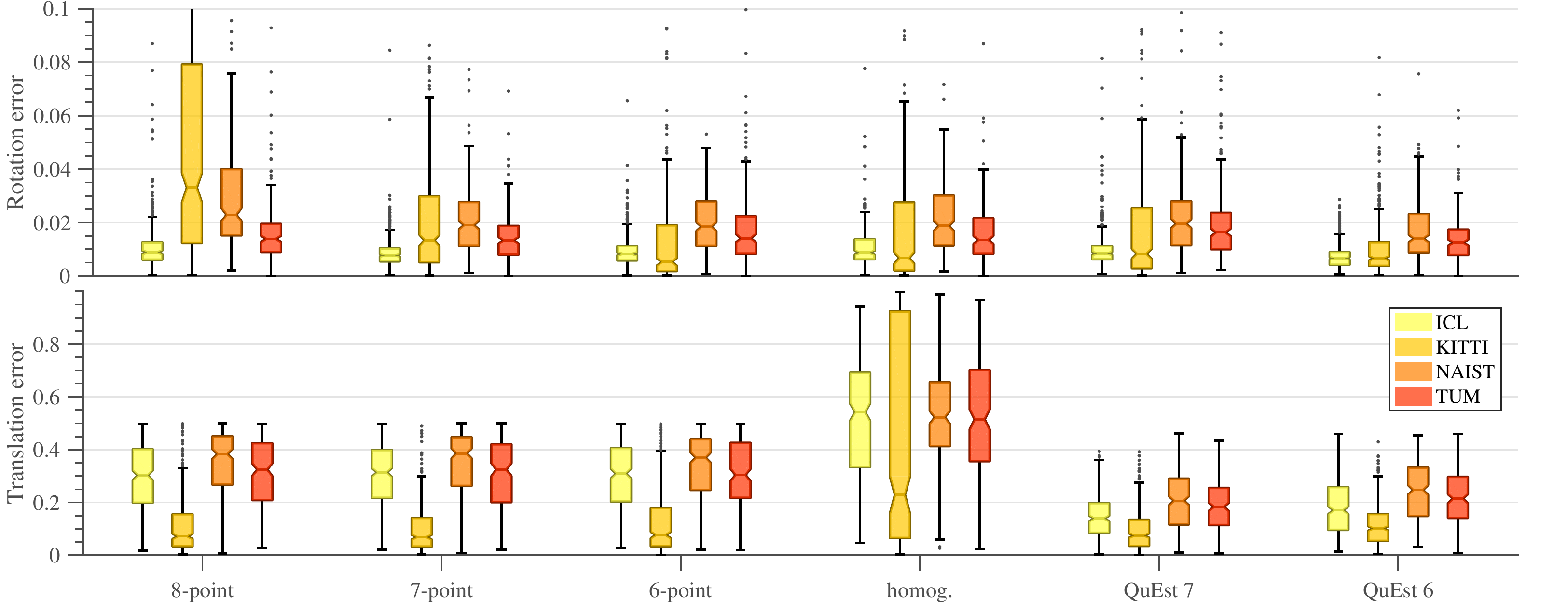}					
		\end{subfigure}
	\end{center}
\caption{Rotation and translation estimation error associated to six algorithms tested on image datasets ICL, KITTI, NAIST, and TUM.}
\label{fig:RotTranBench}
\end{figure*}

The ICL-NUIM dataset consists of computer-generated renderings of 3D
scenes. These artificial images are accompanied by exact ground truth
information, and their main purpose is to benchmark 3D reconstruction
algorithms.
The KITTI and TUM datasets were created to evaluate SLAM algorithms. The
KITTI dataset consists of outdoor stereo image sequences taken from a
moving vehicle. The images are accompanied by LIDAR, IMU and GPS
measurements, so full pose information is provided in the left camera
frame. The TUM dataset is comprised of indoor images as well as point
depth information from a RGB-D camera. The RGB-D camera poses were
recorded by using an optical tracking system, with millimeter-level
accuracy.
The NAIST campus sequences are part of an Augmented Reality benchmark
called TrakMark. The ground truth files for these sequences were created
by solving a PnP problem from known 3D world point coordinates, acquired
using high precision surveying equipment. The images come from a
handheld camera, with the operator walking while capturing some
sequences and running for others.

Figure \ref{fig:RotTranBench} shows the rotation and translation estimation errors associated to each algorithm for all datasets.
As shown in the figure, most algorithms perform well on the ICL benchmark,
due to its lack of image noise. The translation error for all algorithms
is the lowest for the KITTI sequences. This can be explained
by the camera motion, since most of the time the camera is moving
forward, in the direction of its z-axis.
The sequences labeled as NAIST were challenging for every algorithm,
since the camera motion is erratic at times and changes quickly. The TUM
sequences, while generated by a handheld sensor, do not contain motions
as abrupt as those in NAIST. The performance of all algorithms on the
TUM sequences is better than for the NAIST sequences, but below their
performance on the ICL image sets.

Since the chosen feature points may not be coplanar, the homography algorithm does not return the correct solution in general. The pose estimated by the 6-point and 7-point algorithms has smaller median error compared to the 8-point algorithm. 
The translation estimated by QuEst 7 has the smallest median error due to the additional point. The rotation estimated by Quest 6 shows the best performance due to having smaller median errors,  0.25 and 0.75 quartiles, and outliers for all datasets.

\section{Conclusion and Future Work}
\label{sec:conclusion}

By using quaternion representation of rotation, we formulated the camera pose estimation problem and presented the QuEst algorithm to recover the relative pose between two camera views. Unlike the existing homography or essential matrix based methods, QuEst decouples the rotation and translation estimation, and recovers the pose correctly for both cases of general and coplanar points. QuEst can be used to initialize the bundle adjustment algorithm in applications such as SLAM without needing to resort to heuristic methods to detect the coplanarity of the points. Using both simulated and real world images, we demonstrated the estimation accuracy and robustness of QuEst in comparison to the commonly used algorithms.  We have made the Matlab implementation of QuEst available online and free. Future work includes the online release of the C++ implementation of QuEst with RANSAC.

\bibliographystyle{IEEEtran}
\bibliography{msBibs}

\end{document}